\documentclass[10pt,twocolumn,letterpaper]{article}

\usepackage{iccv}
\usepackage{times}
\usepackage{epsfig}
\usepackage{graphicx}
\usepackage{amsmath}
\usepackage{amssymb}

\usepackage{booktabs}
\usepackage{bbding}
\usepackage{makecell}
\usepackage[caption=false,font=footnotesize]{subfig}
\usepackage{float}
\usepackage{balance}
\usepackage[breaklinks=true,bookmarks=false]{hyperref}

\iccvfinalcopy 

\def\iccvPaperID{2767} 

\ificcvfinal\fi

\begin{document}

\title{Learning Dynamic Interpolation for Extremely Sparse Light Fields with Wide Baselines}

\author{Mantang Guo \footnotemark[1]\and
Jing Jin \footnotemark[1]\and
Hui Liu\and
Junhui Hou\and 
Department of Computer Science, City University of Hong Kong, Hong Kong SAR\\
{\tt\small\{mantanguo2-c,jingjin25-c,hliu99-c\}@my.cityu.edu.hk, jh.hou@cityu.edu.hk}
}

\maketitle
\ificcvfinal\thispagestyle{empty}\fi

\renewcommand{\thefootnote}{\fnsymbol{footnote}}
\footnotetext{This work was supported by the Hong Kong RGC under grants CityU 21211518 and 11218121. Corresponding author: Junhui Hou}
\footnotetext[1]{Equal Contributions}

\begin{abstract}
In this paper, we tackle the problem of dense light field (LF) reconstruction from sparsely-sampled ones with wide baselines 
and propose a learnable model, namely dynamic interpolation, to replace the commonly-used geometry warping operation.
Specifically, with the estimated geometric relation between input views, we first construct a lightweight neural network to
dynamically learn weights for interpolating neighbouring pixels from input views to synthesize each pixel of novel views independently. 
In contrast to 
the fixed and content-independent weights employed in the geometry 
warping operation, the learned interpolation 
weights implicitly incorporate the correspondences between the source and novel views and adapt to different image content information.
Then, we recover the spatial correlation between the independently synthesized pixels of each novel view by referring to that of input views using a geometry-based spatial  refinement module. We also 
constrain the angular correlation between the novel views through a disparity-oriented LF structure loss.
Experimental results on LF datasets with wide baselines show that the reconstructed LFs achieve much higher PSNR/SSIM and preserve the LF parallax structure better than  state-of-the-art methods. The source code is publicly available at \url{https://github.com/MantangGuo/DI4SLF}.

\end{abstract}

\section{Introduction}

\begin{figure}[t]
\begin{center}
  \includegraphics[width=0.9\linewidth]{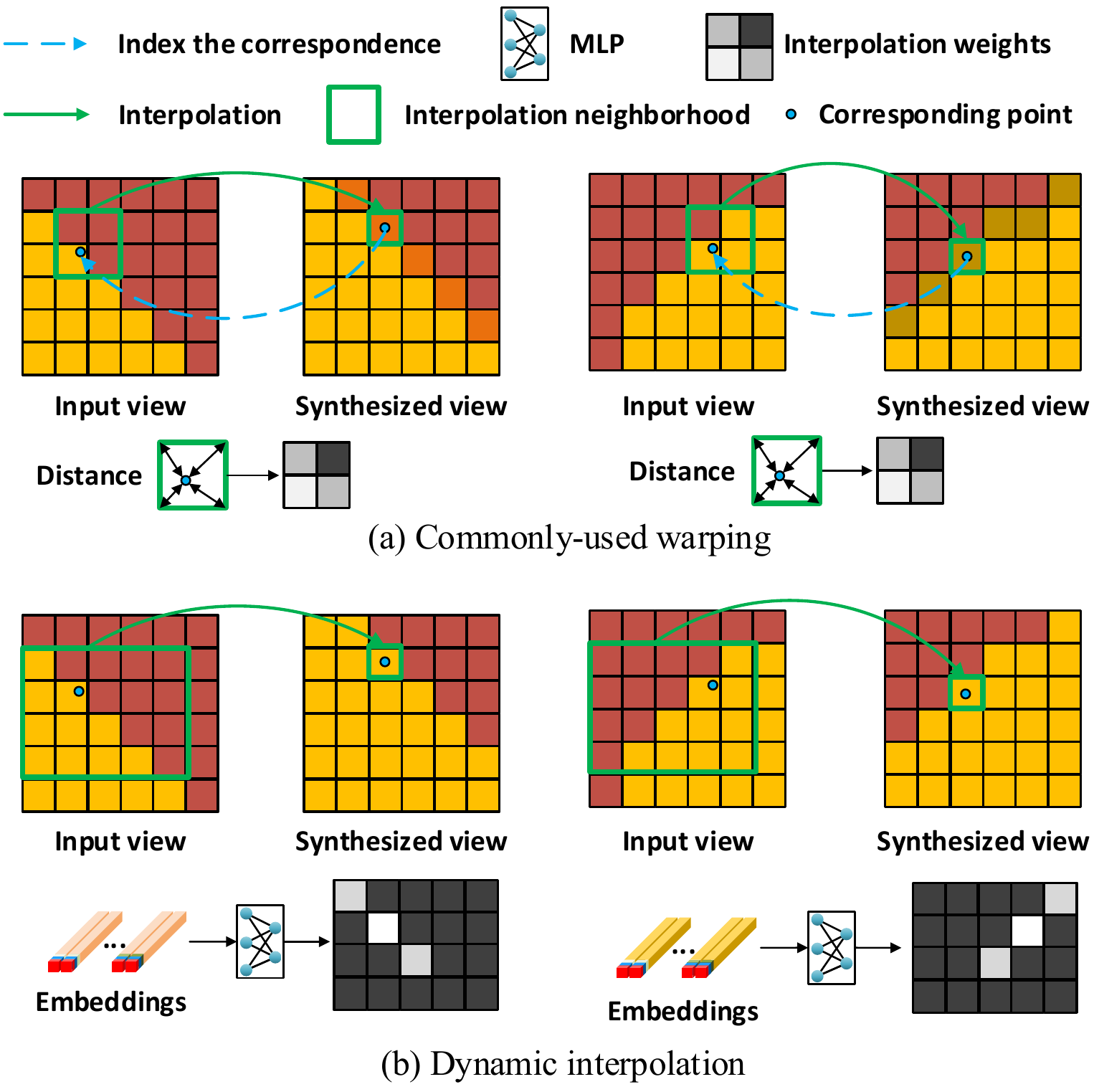}
\end{center}\vspace{-0.4cm}
  \caption{Comparison of the commonly-used warping operation and the proposed dynamic interpolation.
  In contrast to the fixed and content-independent weights employed in the warping operation
  (taking bilinear interpolation weights as an example),
  we propose to dynamically learn geometry-aware and content-adaptive interpolation weights from carefully constructed embeddings.
  }
  \vspace{-0.5cm}
\label{fig:idea}
\end{figure}

Densely-sampled light field (LF) images record not only appearance but also geometry information of 3D scenes, which enable wide applications, such as 3D reconstruction \cite{wang2016depth,Shin_2018_CVPR, guo2020accurate}, image post-refocusing \cite{ng2006digital}, 
and virtual reality \cite{lfapp2015vr,lfapp2017vryu}. 
However, densely-sampled LFs raise great challenges for the acquisition.
For example, camera array \cite{wilburn2005high} or computer-controlled gantry \cite{lfgantry} are either bulky and expensive or limited in capturing static scenes,   
while cost-effective commercial LF cameras \cite{Lytro,RayTrix} suffer from a trade-off between the spatial and angular resolution due to the limited sensor resolution \cite{jin2020light,jin2020hybrid}.

Although many computational methods have been proposed to reconstruct densely-sampled LFs from sparsely-sampled ones, the wide baseline between input views remains a great challenge.
To be more specific, non-depth-based methods \cite{shi2014light,yoon2015learning,wu2017light,wang2018end,yeung2018fast,guo2020deep,guo2021deep} investigate the implicit signal distribution of LF data to learn the mapping from sparse to dense LFs.
These methods inevitably suffer from the aliasing problem and lead to artifacts when the LF is extremely under-sampled.
In comparison, depth-based methods \cite{wanner2013variational, kalantari2016learning,wu2019learning,jin2020learning,jin2020deep} 
perform much better by employing the explicit geometry information.
These methods follow the general pipeline of warping-based view synthesis, and mainly focus on improving the disparity estimation and post-processing refinement.
However, the reconstruction  quality is still limited.


In this paper, we tackle the challenging problem of LF reconstruction from extremely sparse and wide-baseline inputs,
based on an insight that commonly-used warping operation confronts with natural limitations.
Specifically, the warping operation 
synthesizes pixels of the novel view by performing interpolation using their neighboring pixels from input views. 
The employed interpolation weights are determined by fitting a simple and smooth curve using a small set of neighbors,
which inevitably impacts the reconstruction quality as the content information is not considered.
To this end, we propose a learnable module, namely \textbf{dynamic interpolation}, to replace the commonly-used warping operation. As shown in Fig. \ref{fig:idea}, dynamic interpolation uses a lightweight neural network to dynamically predict \textbf{geometry-aware} and \textbf{content-adaptive} interpolation weights for novel view synthesis.
As the pixels of the novel views are independently synthesized,
we subsequently recover the spatial correlation between them by referring to that of input views using a geometry-based refinement module. 
We also constrain the angular correlation between the novel views through a disparity-oriented LF structure loss.
Extensive experimental results demonstrate the significant superiority of the proposed model on  LF datasets with wide baselines over 
warping-based methods as well as other state-of-the-art ones.

In summary, the main contributions of this paper are as follows:\vspace{-0.25cm}
\begin{itemize}
    \item we deeply analyze the geometry warping operation for handling the challenge of LF reconstruction from wide-baseline inputs,  
    and figure out the essential limitation lies in the weakness of the interpolation weights; and
    
    \vspace{-0.25cm}
    \item we reformulate the LF reconstruction from a new perspective and propose dynamic interpolation, which is capable of overcoming the limitation of the geometry warping operation. 
\end{itemize}

\section{Related Work}
The existing LF reconstruction methods could be roughly divided into two categories: non-learning-based methods and learning-based methods.

Non-learning-based methods usually adopt  various prior assumptions to regularize the LF data, i.e., 
Gaussian-based priors \cite{levin2008understanding, levin2010linear, mitra2012light}, sparse priors \cite{marwah2013compressive, shi2014light, vagharshakyan2017light}, and low-rank \cite{kamal2016tensor}. These methods either require many sparse samplings, or have high computational complexity. Explicitly estimating the scene depth information, and then using it to warp input sub-aperture images (SAIs) to novel ones is another kind of methods for LF reconstruction. Wanner and Goldluecke \cite{wanner2013variational} estimated disparity maps at input view by calculating the structure tensor of epipolar plane images (EPIs), and then used the estimated disparity maps to warp input SAIs to the novel viewpoints. This method makes the reconstruction quality rely heavily on the accuracy of the depth estimation. Zhang \textit{et al.} \cite{zhang2015light} proposed a disparity-assisted phase-based method that can iteratively refine the disparity map to minimize the phase difference between the warped novel SAI and the input SAI. However, the angular positions of synthesized SAIs are restricted to the neighbor of input views, which cannot reconstruct LFs with large baselines.

Recently, many deep learning-based methods have been proposed to reconstruct dense LFs from sparse samplings. Yoon \textit{et al.} \cite{yoon2015learning} reconstructed novel SAIs from spatially up-sampled horizontal, vertical and surrounding SAI-pairs by using three separate networks. This method can only regress novel SAIs from adjacent ones, and could not process sparse LFs with large disparities. Wu \textit{et al.} \cite{wu2017light} used a 2-D image super-resolution network to recover high-frequency details along the angular dimension of the interpolated EPI. Analogously, Wang \textit{et al.} \cite{wang2018end} restored the high-frequency details of EPI stacks by using 3-D convolutional neural networks (CNNs). These methods process 2-D or 3-D slices of a 4-D sparse LF, which cannot fully explore the spatial-angular correlations implied in the LF. Yeung \textit{et al.} \cite{yeung2018fast} proposed the computational efficient spatial-angular separable convolution for reconstructing a dense LF from a sparse one in a single forward pass. 

The pipeline of warping-based 
non-learning LF reconstruction methods is also employed by several deep learning-based ones. Kalantari \textit{et al.} \cite{kalantari2016learning} used two sequential networks to separately estimate the disparity map at the novel view, and predicted the color of novel SAI from warped images, respectively. Wu \textit{et al.} \cite{wu2019learning} extracted depth information from the sheared EPI volume, and then used it to reconstruct high angular-resolution EPIs. These methods either ignore the angular relations between synthesized SAIs, or underuse the spatial information of the input SAIs during the reconstruction. Srinivasan \textit{et al.} \cite{srinivasan2017learning} reconstructed an LF from a single 2-D image with predicting 4-D ray depths. This method only works on dataset with small disparities, and is restricted by its generalization ability. Jin \textit{et al.} \cite{jin2020deep} explicitly learned the disparity map at the novel viewpoint from input SAIs. They synthesized the coarse novel SAIs individually by fusing the warped input SAIs with confidence maps. Then they used a refinement network to recover the parallax structure by exploring the complementary information from the coarse LF. Zhou \textit{et al.} \cite{zhou2018stereo} predicted the multiplane image at a reference view by using a CNN to represent the scene's content. Then the novel view can be synthesized from the multiplane image representation with homography and alpha compositing. 

\begin{figure*}[t]
\begin{center} 
  \includegraphics[width=0.88\linewidth]{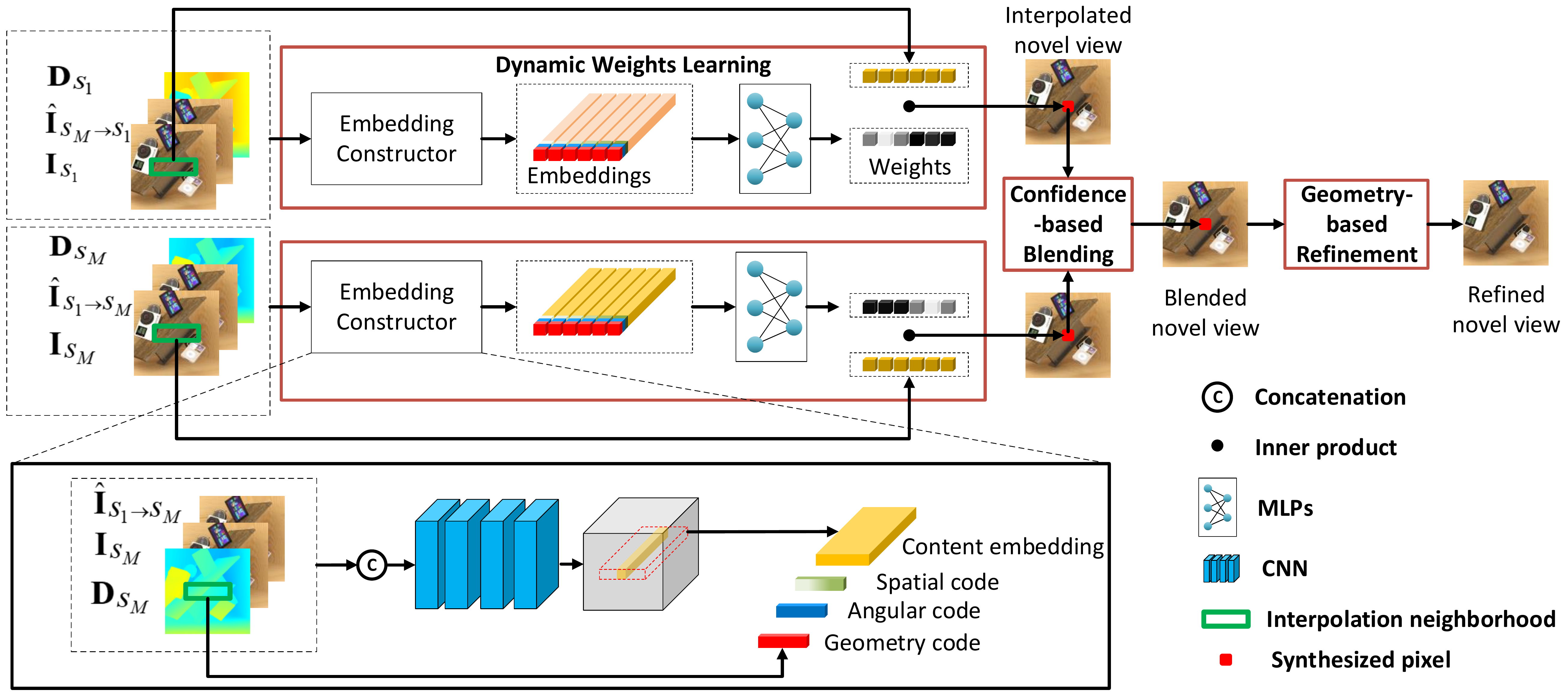}
\end{center}\vspace{-0.4cm}
  \caption{
The flowchart of the proposed \textit{dynamic interpolation} model
for LF reconstruction from extremely sparse (taking $M = 2$ as an example) and wide-baseline inputs. Our proposed model consists of three components: dynamic weight learning, confidence-based blending, and geometry-based spatial refinement.
}
\label{fig:pipeline}\vspace{-0.4cm}
\end{figure*}

\section{Problem Analysis}





Denote by $\mathcal{L}(u,x,y)\in \mathbb{R}^{U \times H\times W}$ a 3-D LF containing $U$ sub-aperture images (SAIs) of each  spatial resolution $H\times W$, which are sampled along a 1-D straight line.
The SAI at the angular position of $u$ is denoted as $\mathbf{I}_u$.
Given an extremely sparse LF with $M$ SAIs, denoted by $\mathcal{S}=\{\mathbf{I}_{s_1},\cdots,\mathbf{I}_{s_M}\}$,  where $M\ll U$, 
our goal is to synthesize the unsampled SAIs, denoted by $\mathcal{T} =\{\mathbf{I}_{t_1},\cdots,\mathbf{I}_{t_N}\}$,  where $N=U-M$,
so that the densely-sampled LF denoted by $\widetilde{\mathcal{L}}$ can be reconstructed as close to $\mathcal{L}$ as possible.
This problem can be implicitly formulated as:\vspace{-0.2cm}
\begin{equation}
 \widetilde{\mathcal{L}}= \mathcal{S}\bigcup\widetilde{\mathcal{T}}  = f(\mathcal{S}), \vspace{-0.25cm}
\end{equation}
where $\widetilde{\mathcal{T}}$ is the set of synthesized novel SAIs, and $f$ denotes the mapping function to be learned.
Not that we also denote the input and target SAI as $\mathbf{I}_s$ and $\mathbf{I}_t$, respectively, in the rest of the paper for simplification.
The SAIs of a 3-D LF image are the observations of the same scene from different viewpoints.
Under the assumption of Lambertian and non-occlusion, projections of the same scene point will have the same intensity at different SAIs.
This relation can be described as:\vspace{-0.25cm}
\begin{equation}
    \label{eq:LFstructure}
    \mathbf{I}_{t}(\mathbf{x}_t) = \mathbf{I}_{s}(\mathbf{x}'_t),
    \vspace{-0.25cm}
\end{equation}
where $\mathbf{x}_t=(x_t,y_t)$ is the spatial coordinate,
and $\mathbf{x}'_t$ is the location of the corresponding pixel of $\mathbf{I}_{t}(\mathbf{x}_t)$ at $\mathbf{I}_{s}$.
Given the disparity value of $\mathbf{I}_{t}(\mathbf{x}_t)$, denoted by $d$, $\mathbf{x}'_t$ can be easily computed as $\mathbf{x}'_t = (x'_t,y'_t)=(x_t+d(s-t),y_t)$.
Based on this relation, the pixels of $\mathbf{I}_t$ can be estimated by collecting their corresponding pixels on $\mathbf{I}_s$.
However, as the values of $\mathbf{x}'_t$ are always fractional, interpolation is required to computed the intensity of the corresponding pixel by the weighted sum of the neighboring pixels.
The interpolation process can be formulated as:\vspace{-0.2cm}
\begin{equation}
    \mathbf{I}_{t}(\mathbf{x}_t) = \sum_{\mathbf{x}_s\in\mathcal{P}_{\mathbf{x}'_t}} w(\mathbf{x}_s-\mathbf{x}'_t ; \phi_w ) \mathbf{I}_s (\mathbf{x}_s),
    \vspace{-0.25cm}
\end{equation}
where 
$\mathcal{P}_{\mathbf{x}'_t}$ is the set of  neighbors of $\mathbf{x}'_t$, and $w$ is the function with the parameter $\phi_w$ which defines the interpolation weights for the pixels of $\mathbf{I}_s$.


The above mentioned procedure is adopted by the commonly-used warping operation.
However, we provide an insight that this procedure has natural limitations from two aspects:

(1)
It requires to estimate $d$ to locate $\mathbf{x}'_t$. However, estimating the disparities of unsampled SAIs from the input SAIs is challenging.
Moreover, as $\mathcal{P}_{\mathbf{x}'_t}$ is always a small set of pixels surrounding $\mathbf{x}'_t$, e.g., $2$ (or $2\times2$) neighbors for linear (or bilinear) interpolation \cite{srinivasan2017learning} and $4$ (or $4\times4$) neighbors for cubic (or bicubic) interpolation \cite{kalantari2016learning},
the reconstruction results greatly rely on the accuracy of the disparity estimation.

(2) The weight function $w(\cdot;\phi_w)$ is defined by fitting a simple and smooth curve using the small set $\mathcal{P}_{\mathbf{x}'_t}$,
which neglects the content information.
Thus, even with an accurate estimation of $d$, it is difficult to produce high-quality results, especially on areas with texture edges, occlusion boundaries, and non-Lambertian objects.

Base on these observations, we propose a novel model, namely \textit{dynamic interpolation}, to synthesize $\widetilde{\mathcal{T}}$ from $\mathcal{S}$,
which overcomes the limitations of the commonly-used warping operation by learning geometry-aware and content-adaptive interpolation weights.

\section{Proposed Method}




\textbf{Overview}. As shown in Fig.~\ref{fig:pipeline}, the proposed model, namely \textit{dynamic interpolation}, mainly consists of three components, i.e., dynamic weight learning, confidence-based blending, and geometry-based spatial refinement.
Specifically, 
we first independently synthesize each pixel of $\widetilde{\mathbf{I}}_t$  by applying interpolation over its neighboring pixels from $\mathbf{I}_s$.
In contrast to commonly-used warping operation, the interpolation weights used in our model are dynamically learned. 
We also estimate confidence maps to blend the pixels interpolated from different input SAIs, which further handles the occlusion problems.
Then, we recover the spatial correlation between the independently synthesized pixels by referring to that of $\mathbf{I}_s$.

In this paper, we set $M=2$.
It is worth noting that our framework could be straightforwardly extended to 4-D LFs with larger $M$. 
In what follows, we will introduce the technical details of the proposed method. 
 


\subsection{Dynamic Weight Learning}



This module aims at learning the interpolation weights to independently synthesize each pixel of a novel SAI, denoted by $\mathbf{I}_t(\mathbf{x}_t)$, from $\mathbf{I}_s$.
The interpolation weight for each pixel $\mathbf{I}_s(\mathbf{x}_s)$ is predicted by a multilayer perceptron (MLP), and
the following information is embedded in the MLP:

(1) \textbf{The correspondence relation} between $\mathbf{I}_t$ and $\mathbf{I}_s$, which helps to implicitly locate the corresponding pixel of $\mathbf{I}_t(\mathbf{x_t})$ in $\mathbf{I}_s$.
The correspondence embedding consists of three components, i.e., a geometric code $E_{\mathbf{x}_t,\mathbf{x}_s}^{geo}$, a spatial code $\mathbf{E}_{\mathbf{x}_t,\mathbf{x}_s}^{spa}$, and an angular code $E_{\mathbf{x}_t,\mathbf{x}_s}^{ang}$.
Specifically, 
$E_{\mathbf{x}_t,\mathbf{x}_s}^{geo}$ is the disparity value of $\mathbf{I}_s(\mathbf{x}_s)$, i.e., \vspace{-0.2cm}
\begin{equation}
    \begin{aligned}
        E_{\mathbf{x}_t,\mathbf{x}_s}^{geo} = \mathbf{D}_s(\mathbf{x}_s),
    \end{aligned}
     \vspace{-0.2cm}
\end{equation}
where $\mathbf{D}_s$ is the disparity map of $\mathbf{I}_s$, which is estimated from $\mathcal{S}$ using a pre-trained optical-flow model.
$E_{\mathbf{x}_t,\mathbf{x}_s}^{spa}$ and $E_{\mathbf{x}_t,\mathbf{x}_s}^{ang}$ describe the the spatial and angular distance between $\mathbf{I}_t(\mathbf{x}_t)$ and $\mathbf{I}_s(\mathbf{x}_s)$, i.e.,\vspace{-0.2cm}
\begin{equation}
    \begin{aligned}
        \mathbf{E}_{\mathbf{x}_t,\mathbf{x}_s}^{spa} &= \mathbf{x}_s - \mathbf{x}_t,\
        E_{\mathbf{x}_t,\mathbf{x}_s}^{ang} &= s - t.
    \end{aligned}
    \vspace{-0.2cm}
\end{equation}
These information can directly determine whether $\mathbf{I}_s(\mathbf{x}_s)$ corresponds to $\mathbf{I}_t(\mathbf{x}_t)$ under the estimated geometric relation,
and thus, greatly helps the MLP to locate informative pixels in $\mathbf{I}_s$ and allocate larger weights to them.

(2)\textbf{ The content information} around $\mathbf{I}_s(\mathbf{x}_s)$, which helps to understand complicated scenarios, such as texture edges, occlusion boundaries, and non-Lambertian objects.
To construct the content embedding, denoted by $\mathbf{E}_{\mathbf{x}_t,\mathbf{x}_s}^{ctt}$,
we first backward warp the other SAI in $\mathcal{S}$ to $\mathbf{I}_s$ based on $\mathbf{D}_s$, and the resulting image is denoted by $\widehat{\mathbf{I}}_{s'\rightarrow s}$.
Then, we employ a sub-CNN $f_c(\cdot)$ to learn the content information, i.e.,
\vspace{-0.1cm}
\begin{equation}
    \begin{aligned}
        \mathbf{E}_{\mathbf{x}_t,\mathbf{x}_s}^{ctt} &= f_c\left(\mathbf{x}_s, \mathbf{x}_t,\mathbf{I}_s, \widehat{\mathbf{I}}_{s'\rightarrow s}, \mathbf{D}_s\right).
    \end{aligned}
\vspace{-0.1cm}
\end{equation}
It is expected that $f_c$ is able to detect the texture edges of $\mathbf{I}_s$ and  
understand the occlusion and non-Lambertian relations by comparing $\mathbf{I}_s$ and  $\widehat{\mathbf{I}}_{s'\rightarrow s}$ with the assistance of $\mathbf{D}_s$.

Finally, the geometry and content embedding, denoted by
$\mathbf{E}_{\mathbf{x}_t,\mathbf{x}_s}$,
is constructed as:
\vspace{-0.1cm}
\begin{equation}
    \begin{aligned}
        \mathbf{E}_{\mathbf{x}_t,\mathbf{x}_s} = \textsf{CAT}\left(
        E_{\mathbf{x}_t,\mathbf{x}_s}^{geo}, \mathbf{E}_{\mathbf{x}_t,\mathbf{x}_s}^{spa},
        E_{\mathbf{x}_t,\mathbf{x}_s}^{ang},
        \mathbf{E}_{\mathbf{x}_t,\mathbf{x}_s}^{ctt} \right),
    \end{aligned}
    \vspace{-0.1cm}
\end{equation}
where $\textsf{CAT}(\cdot)$ is the concatenation operation,
and the interpolation weights for $\mathbf{I}_s(\mathbf{x}_s)$ to synthesize $\mathbf{I}_t(\mathbf{x}_t)$, denoted by $W_{\mathbf{x}_t,\mathbf{x}_s}$, is predicted as:
\vspace{-0.1cm}
\begin{equation}
    \begin{aligned}
        W_{\mathbf{x}_t,\mathbf{x}_s} = f_w\left(\mathbf{E}_{\mathbf{x}_t,\mathbf{x}_s}\right),
    \end{aligned}
\vspace{-0.1cm}
\end{equation}
where $f_w(\cdot)$ is the learnable MLP.

To reduce the computational cost, the interpolation is performed over the neighborhood of $\mathbf{I}_t(\mathbf{x}_t)$ in $\mathbf{I}_s$, instead of the whole range of $\mathbf{I}_s$.
Suppose the disparity range of $\mathbf{I}_t$  is $[-d_{max},d_{max}]$, the neighborhood of $\mathbf{I}_t(\mathbf{x}_t)$ in $\mathbf{I}_s$ is defined as:
$\mathcal{P}_{\mathbf{x}_t} = \{\mathbf{x}=(x,y)|x_t-d_{max}(s-t) \leq x \leq x_t+d_{max}(s-t), y = y_t\}$.
Then, 
we predict the intensity of $\mathbf{I}_t(\mathbf{x}_t)$ by
applying interpolation on $\mathcal{P}_{\mathbf{x}_t}$ based on the learned weights, and the predicted result is denoted by $\widetilde{\mathbf{I}}_{s\rightarrow t} (\mathbf{x}_t)$, i.e.,
\vspace{-0.2cm}
\begin{equation}
    \begin{aligned}
        \widetilde{\mathbf{I}}_{s\rightarrow t} (\mathbf{x}_t) = \sum_{\mathbf{x}_s\in\mathcal{P}_{\mathbf{x}_t}} W_{\mathbf{x}_t,\mathbf{x}_s} \mathbf{I}_s(\mathbf{x}_s).
    \end{aligned}
    \vspace{-0.2cm}
\end{equation}

\textit{Remark}:
Compared with the commonly-used warping operation, our dynamic interpolation has the following advantages:

(1)  
Instead of relying on the disparity estimation accuracy of the novel SAI, we utilize the disparity map directly estimated between input SAIs, which is much more reliable.
Moreover, the geometry information is implicitly incorporated by learning the weights for each pixel in the possibly maximum neighborhood, which might improve the tolerance of the disparity estimation error.

(2) Instead of fitting a simple curve using a small set of pixels for interpolation,
we learn the weights using an MLP 
and provide content information over a relatively large field  to make the weights adaptive to various and complicated neighboring correlations.


\subsection{Confidence-based Blending} 
Although the weight learning module has the ability of handling the problem of occlusion boundaries by embedding the content information,
it is still difficult to synthesize the pixels whose correspondences are completely occluded in $\mathbf{I}_s$ by interpolation on only one of the input SAIs.
Fortunately, the object occluded from one viewpoint might be visible from another one.
Therefore, we blend the images synthesized from different input SAIs
under the guidance of their confidence maps, which indicate the non-occlusion pixels with higher values.

To predict the confidence value for each pixel position $\mathbf{x}_t$ in the synthesized SAI,
we first aggregate the geometry and content embeddings for each neighbors of $\mathbf{x}_t$ in $\mathbf{I}_s$ by concatenation, and then apply another MLP, denoted by $f_b(\cdot)$, on the aggregated feature, i.e., \vspace{-0.1cm}
\begin{equation}
\begin{aligned}
\widetilde{\mathbf{C}}_{s\rightarrow t}(\mathbf{x_t}) = f_b\left(\textsf{CAT}\{\mathbf{E}_{\mathbf{x}_t,\mathbf{x}_s}|\mathbf{x}_s\in\mathcal{P}_{\mathbf{x}_t}\}\right),
\end{aligned}
\vspace{-0.1cm}
\end{equation}
where $\widetilde{\mathbf{C}}_{s\rightarrow t}$ is the confidence map for $\widetilde{\mathbf{I}}_{s\rightarrow t}$.
Based on the learned confidence map, the SAIs synthesized from different input SAIs are combined to produce the intermediate result of the novel SAI, denoted by $\widetilde{\mathbf{I}}_t^b$,
i.e.,
\begin{equation}
\begin{aligned}
\widetilde{\mathbf{I}}_t^b = \sum_{s\in\{s_1,\cdots,s_M\}} \widetilde{\mathbf{C}}_{s\rightarrow t}\odot \widetilde{\mathbf{I}}_{s\rightarrow t},
\end{aligned}
\vspace{-0.3cm}
\end{equation}
where $\odot$ is the element-wise multiplication operator.





\subsection{Geometry-based Spatial Refinement}

As pixels in $\widetilde{\mathbf{I}}_t^b$ are independently synthesized, the spatial correlations among them are not considered.
To further enhance the quality of $\widetilde{\mathbf{I}}_t^b$,
we propose a refinement module to recover its spatial correlation by inferring that from the input SAIs using a sub-CNN.
Considering the wide baseline between $\widetilde{\mathbf{I}}_t^b$ and $\mathbf{I}_s$, directly applying a network will have difficulties to perceive the corresponding information from $\mathbf{I}_s$.
Therefore, we adopt a geometry-based spatial refinement, which first explicitly locates the correspondences in $\mathbf{I}_s$ at the patch level, and then learn the spatial correlations from $\mathbf{I}_s$ to refine $\widetilde{\mathbf{I}}_t^b$.


Let $\widetilde{\mathbf{H}}_t^{\mathbf{x}_t^o}$ denote a patch of $\widetilde{\mathbf{I}}_t^b$ centered at $\mathbf{x}_t^o=(x_t^o,y_t^o)$.
To locate its similar patch in $\mathbf{I}_s$, we first estimate the disparity map of $\widetilde{\mathbf{I}}_t^b$ by forward warping $\mathbf{D}_s$, resulting in $\widetilde{\mathbf{D}}_t$, and then calculate the patch-level disparity of $\widetilde{\mathbf{H}}_t^{\mathbf{x}_t^o}$ by averaging the disparity over all of its contained pixels, leading to  $\widetilde{d}_h$.
Then, the central position of the corresponding patch of $\widetilde{\mathbf{H}}_t^{\mathbf{x}_t^o}$ at $\mathbf{I}_s$, denoted by $\mathbf{x}_s^o=(x_s^o,y_s^o)$, can be estimated as:
\vspace{-0.1cm}
\begin{equation}
\begin{aligned}
x_s^o = x_t^o + \widetilde{d}_h(s-t).
\end{aligned}
\vspace{-0.25cm}
\end{equation}
Based on $\mathbf{x}_s^o$, we can collect the corresponding patch of $\widetilde{\mathbf{H}}_t^{\mathbf{x}_t^o}$ at $\mathbf{I}_s$, which is denoted by $\mathbf{H}_s^{\mathbf{x}_s^o}$.

To recover the spatial correlation among pixels in  $\widetilde{\mathbf{H}}_t^{\mathbf{x}_t^o}$,
we feed the concatenation of $\widetilde{\mathbf{H}}_t^{\mathbf{x}_t^o}$ and its corresponding patches in all the input SAIs, i.e., $\{\mathbf{H}_s^{\mathbf{x}_s^o}|s\in\{s_1,\cdots,s_M\}\}$,
into a sub-CNN to predict a residual map for refinement.
We then merge the refined patches to produce the final prediction of the novel SAI, i.e.,
\vspace{-0.1cm}
\begin{equation}
    \begin{aligned}
        \widetilde{\mathbf{I}}_t = f_r(\widetilde{\mathbf{I}}_t^b),
    \end{aligned}
    \vspace{-0.1cm}
\end{equation}
where $f_r(\cdot)$ denotes the geometry-based spatial refinement module.

\subsection{Disparity-oriented Loss}

The final and intermediate predictions of the novel SAIs are supervised by the ground-truth one, i.e., the loss function for the reconstruction of $\mathbf{I}_t$ is defined as:
\vspace{-0.1cm}
\begin{equation}
\begin{aligned}
        \ell^r_t = \left\|\widetilde{\mathbf{I}}_t-\mathbf{I}_t\right\|_1 
        + \left\|\widetilde{\mathbf{I}}_t^b-\mathbf{I}_t\right\|_1
        + \sum_{s\in\{s_1,\cdots,s_M\}} \left\| \widetilde{\mathbf{I}}_{s\rightarrow t}-\mathbf{I}_t\right\|_1.
\end{aligned}
\vspace{-0.1cm}
\end{equation}

Moreover, as each novel SAI is reconstructed individually, 
we propose a disparity-oriented LF structure loss to constrain the angular correlation between them. 
The relation described in Eq.~(\ref{eq:LFstructure}) 
can be constrained by minimizing the gradients along the directions of the straight lines in EPIs of the LF. Considering the existence of occlusions and non-Lambertian, we instead 
minimizing the distance between the gradients of predicted EPIs and the ground-truth ones.
Note that the directions of EPI lines are located under the guidance of the ground-truth disparity of the LFs, which are easily available in the training datasets. 
Such a disparity-oriented LF structure loss can be formulated as \vspace{-0.15cm}
\begin{equation}
\begin{aligned}
\ell^d = \left\|\nabla_d \widetilde{\mathcal{E}} -\nabla_d \mathcal{E} \right\|_1,
\end{aligned}
\vspace{-0.15cm}
\end{equation}
where $\nabla_d$ is the gradient operator along the direction defined by the ground-truth disparity $d$ of each pixel, and $\widetilde{\mathcal{E}}$ and $\mathcal{E}$ are the EPIs of $\widetilde{\mathcal{L}}$ and $\mathcal{L}$, respectively.

Our framework is end-to-end trained using the final objective function defined as: $\ell = \sum_{t\in\{t_1,\cdots,t_N\}} \ell_t^r  + \lambda \ell^d$, where $\lambda\geq 0$ is the weight factor for the disparity-oriented loss.

\begin{table*}[t]
\centering
\setlength{\tabcolsep}{1.5mm}
\caption{Quantitative comparisons (PSNR/SSIM) of different methods over the Inria Sparse LF dataset \cite{shi2019framework}.}
\begin{tabular}{c|c|cccccc}
\toprule
        Light Field &Disparity range & Baseline  & \makecell{Kalantari\\\textit{et al.} \cite{kalantari2016learning}}  & \makecell{Wu\\\textit{et al.} \cite{wu2019learning}} & \makecell{Jin\\\textit{et al.} \cite{jin2020deep}} &\makecell{Ours\\(PWCNet)} & \makecell{Ours\\(RAFT)}   \\ 
\midrule
Electro\_devices    &[-19.6, 32.8]    &28.49/0.871  &24.66/0.691                                             &28.51/0.866                              &32.77/0.936                       &33.04/0.941  &\textbf{35.43/0.960}\\
Flying\_furniture   &[-34.0, 62.4]   &28.39/0.838  &28.83/0.784                                             &27.38/0.783                              &31.69/0.896                        &30.06/0.881   &\textbf{33.93/0.935}\\
Coffee\_beans\_vases&[10.8, 58.4]    &27.17/0.886  &21.54/0.579                                             &23.04/0.836                              &28.08/0.927                        &\textbf{29.63}/0.936  &29.55/\textbf{0.943}\\
Dinosaur            &[-57.6, 72.8]  &23.00/0.773  &22.21/0.731                                             &23.07/0.788                              &26.61/0.897                         &24.94/0.861 &\textbf{27.50/0.904}\\
Flowers             &[-40.4, 66.0]  &23.05/0.757  &21.96/0.667                                             &23.52/0.767                              &24.36/0.842                         &\textbf{25.24/0.860}  &24.86/0.849\\
Rooster\_clock      &[-34.4, 21.2]    &31.43/0.904  &22.71/0.710                                             &29.05/0.887                              &27.69/0.929                       &35.90/0.946    &\textbf{38.16/0.966}\\
Smiling\_crowd      &[-40.4, 64.8]  &18.87/0.722  &17.01/0.596                                             &19.52/0.729                              &21.01/0.822                         &20.90/0.824  &\textbf{22.87/0.877}\\
\bottomrule
\multicolumn{2}{c|}{Average}              &25.77/0.821  &22.70/0.680                                             &25.05/0.802                              &27.46/0.893                    &28.53/0.893      &\textbf{30.33/0.919}\\
\bottomrule
\end{tabular}
\label{table:quantitative}
\vspace{-0.6cm}
\end{table*}

\section{Experimental Results}

\subsection{Training Details and Datasets}
Both the content embedding network $f_c(\cdot)$ and the spatial refinement network $f_r(\cdot)$ are 2-D CNNs that consist of $4$ residual blocks \cite{he2016deep} with the kernel  of size $3\times3$.
Zeros-padding was applied to keep the spatial size unchanged. 
We refer readers to the \textit{Supplementary Material} for the detailed network architecture.
At each iteration of the training phase,
a $32\times32$ patch randomly cropped from the LF image was synthesized.
We used the \textit{unfold} function in \textit{PyTorch} to efficiently locate the neighborhood of each synthesized pixel.

The batch size was empirically set to $1$. The learning rate was initially set to $1e^{-4}$  and reduced to $1e^{-5}$ after $8000$ epochs. We used Adam \cite{kingma2015adam} with $\beta_1=0.9$ and $\beta_2=0.999$ as the optimizer.

We trained our framework using $29$ LF images from the Inria Sparse LF dataset \cite{shi2019framework}.
Each LF image contains $9\times 9$ SAIs with a disparity range of $[-20,20]$ between adjacent SAIs.
We took the 3$^{rd}$ and 7$^{th}$ SAIs at the same row as inputs, which have a wide baseline up to $80$ pixels, to train the framework. 
The test dataset consists of $7$ LF images from the Inria Sparse LF dataset  \cite{shi2019framework} and $14$ LF images from the MPI LF archive \cite{Vamsi2017}. Note that MPI \cite{Vamsi2017} is a high angular-resolution LF dataset where each LF image contains $101$ SAIs distributed on a scanline. Thus, we can construct testing LFs with different baselines by sampling SAIs with different intervals (see details in Section \ref{sec:exp_comp} ).

\subsection{Comparison with State-of-the-art Methods}
\label{sec:exp_comp}

We compared the proposed method with three state-of-the-art deep learning-based methods for LF reconstruction, including Kalantari \textit{et al.} \cite{kalantari2016learning}, Wu \textit{et al.} \cite{wu2019learning}, and Jin \textit{et al.} \cite{jin2020deep}. 
All the methods were retrained on the same dataset with the officially released codes and suggested configurations.
Note that the 2-D angular convolutional layers used by Jin \textit{et al.} \cite{jin2020deep} were degenerated to 1-D convolutional layers to adapt to the 3-D LF.


To verify the advantage of the proposed dynamic interpolation in comparison to the commonly-used warping operation,
we developed a baseline model by replacing the dynamic weight learning module with a disparity-based warping operation while leaving the confidence-based blending and the geometry-based refinement unchanged.
Note that to ensure fair comparisons,
the disparity maps of the novel views used for warping were estimated from the same inputs as Ours, i.e., the disparity maps of the input views from a pre-trained optical-flow model.


Moreover, to demonstrate the effects of the input disparity estimation accuracy  on the performance of our model, we adopted two different optical-flow methods, namely RAFT \cite{teed2020raft} and PWCNet \cite{Sun_2018_CVPR}, and used them to separately train two models, denoted as  Ours (RAFT) and Ours (PWCNet), respectively.

\begin{figure}[t]
\centering
\subfloat[]{\includegraphics[width=0.255\textwidth]{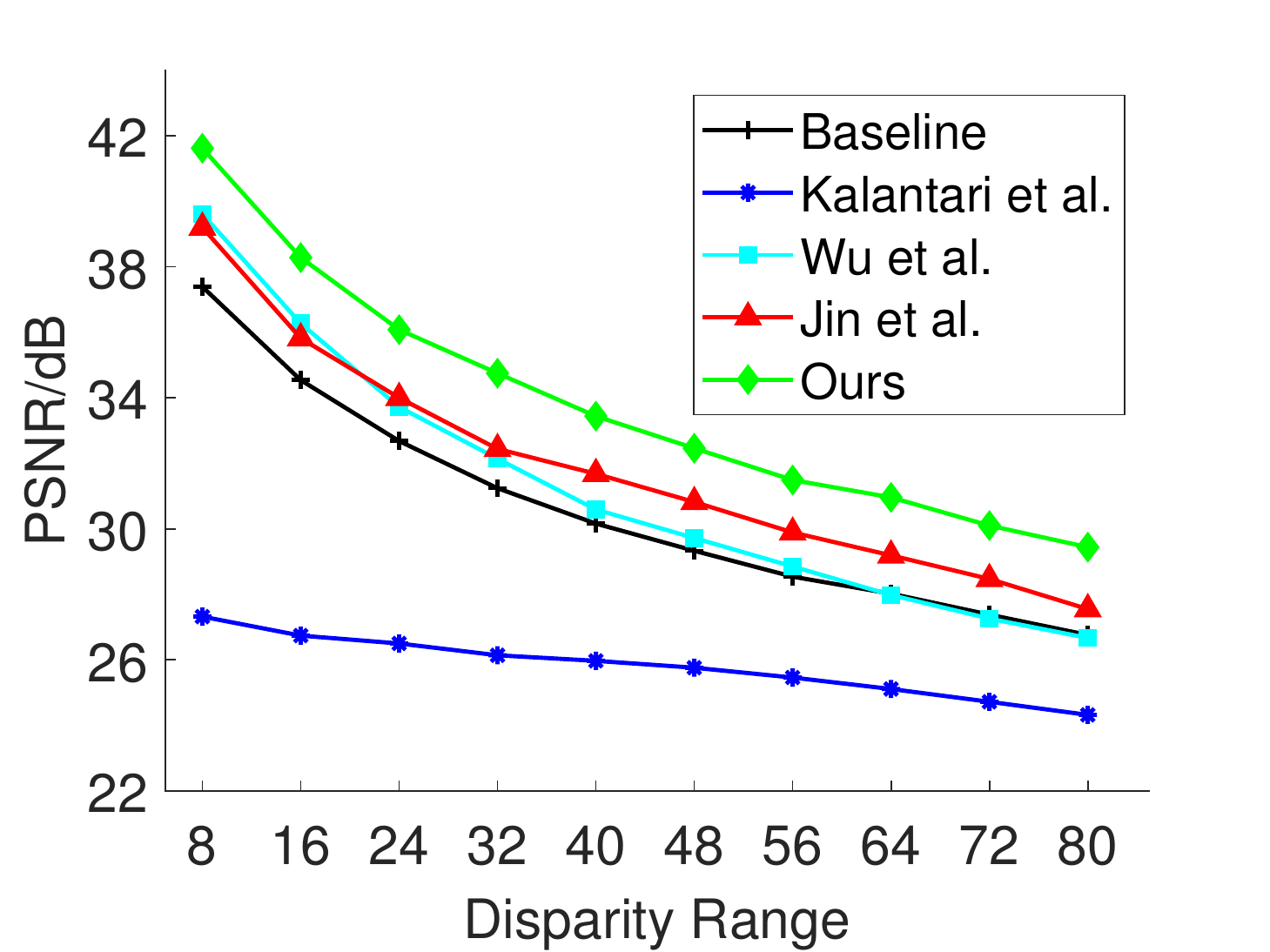}}
\subfloat[]{\includegraphics[width=0.255\textwidth]{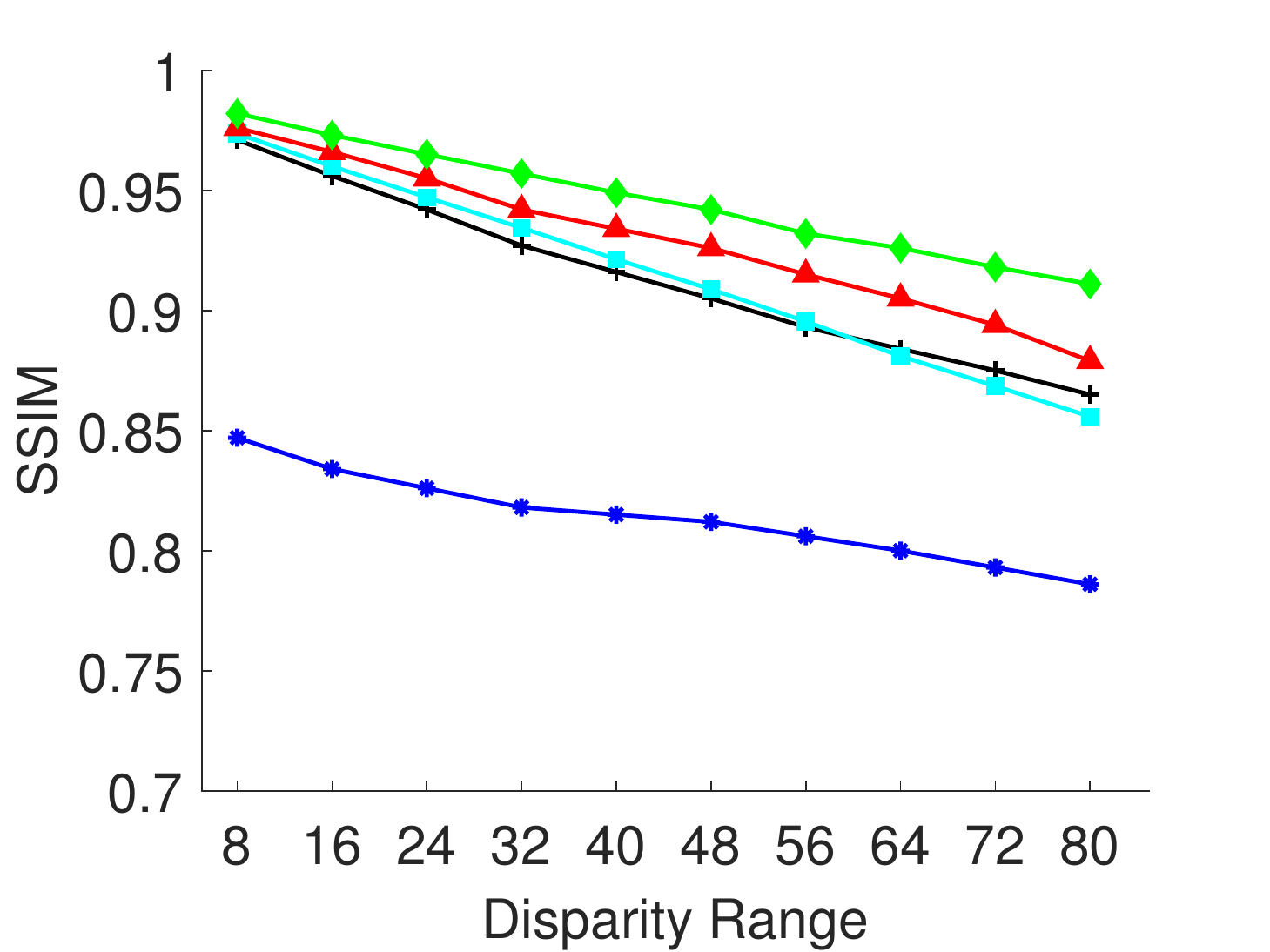}}
\caption{Quantitative comparisons (PSNR/SSIM) of different methods under different disparity ranges (pixels) between input SAIs on  the MPI LF dataset \cite{Vamsi2017}. All subfigures share the same legend shown in the first one.}
\label{fig:quanti_mpi}
\vspace{-0.4cm}
\end{figure}

\begin{figure*}[t]
\begin{center}
  \includegraphics[width=0.85\linewidth]{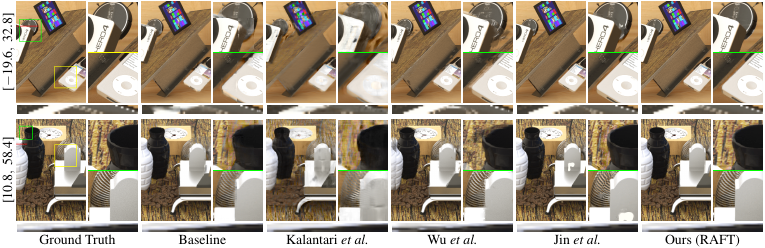}
\end{center}\vspace{-0.4cm}
  \caption{Visual comparisons of reconstructed LFs from different methods over the Inria Sparse LF dataset \cite{shi2019framework}. The disparity range between input SAIs of each LF is shown on the left.} 
\label{fig:visual_inria}
\vspace{-0.3cm}
\end{figure*}

\begin{figure*}[t]
\begin{center}
  \includegraphics[width=0.85\linewidth]{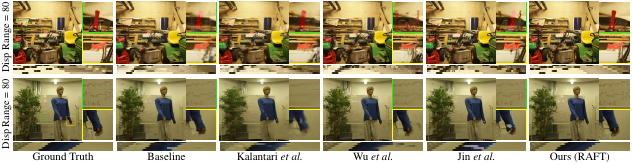}
\end{center}\vspace{-0.4cm}
  \caption{Visual comparisons of reconstructed LFs from different methods over the MPI LF dataset \cite{Vamsi2017}. The disparity range between input SAIs reaches $80$ pixels for each reconstructed LF.}
\label{fig:visual_mpi}
\vspace{-0.5cm}
\end{figure*}

\textbf{Quantitative comparisons on the Inria Sparse LF dataset.}
We reconstructed  $5\times5$ LF images in the Inria dataset \cite{shi2019framework} in a row-by-row manner, and  calculated the PSNR and SSIM values between the reconstructed LFs and ground-truth ones in Y channel to quantitatively evaluate different methods.  Table~\ref{table:quantitative} lists the results, where it can be observed that:\vspace{-0.25cm}
\begin{itemize}
    \item Ours (RAFT) achieves significantly higher PSNR and SSIM than Baseline. Although the disparity maps between input SAIs are also provided to Baseline, its performance is still limited by the fixed and content-independent weights employed in the commonly-used warping operation, which demonstrates the advantage of learned dynamic weights in our method;\vspace{-0.25cm}
    
    \item Ours (PWCNet) is worse than Ours (RAFT) but still much better than all the compared methods. Although the estimated disparity maps with different levels of quality are involved, our method may adaptively aggregate content features to provide gains for learning dynamic weights, which can demonstrate the advantages of learning dynamic weights over commonly-used ones. Besides, it is highly expected that our framework will be further improved with more powerful and advanced optical flow estimation proposed in future;  \vspace{-0.25cm}
    
    \item our method achieves higher performance than Wu \textit{et al.} \cite{wu2019learning}. The reason maybe that Wu \textit{et al.} \cite{wu2019learning} reconstructed LFs in the perspective of reconstructing 2-D EPIs, and neglected modeling the correlations between two spatial dimensions, which limits the performance. On the contrary, our method employs a geometry-based refinement module to refine the correlations among pixels of novel SAIs, and further improves the reconstruction quality;  and
    \vspace{-0.25cm}
    
    \item both Kalantari \textit{et al.} \cite{kalantari2016learning} and Jin \textit{et al.} \cite{jin2020deep} achieve worse performances than our method. In addition to that they cannot estimate disparities well from extremely sparse LFs with a limited receptive field of CNNs, they also cannot handle severe artifacts brought by the warping operation. By contrast, our method can effectively mitigate the warping errors in reconstructing extremely sparse LFs by dynamically learning content-adaptive weights for each pixel of the novel SAI. Besides, the geometry-based refinement module can further refine novel SAIs to improve the quality. 
    
\end{itemize}

\begin{figure*}[thp]
\begin{center}
  \includegraphics[width=0.85\linewidth]{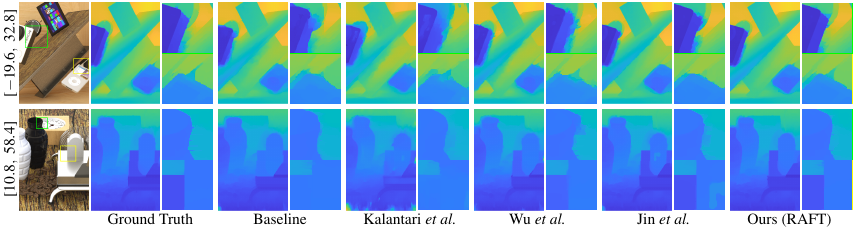}
\end{center}\vspace{-0.4cm}
  \caption{Visual comparisons of estimated depth maps from different methods over the Inria Sparse LF dataset \cite{shi2019framework}. The disparity range between input SAIs of each LF is shown on the left.}
\label{fig:visual_inria_depth}
\vspace{-0.5cm}
\end{figure*}

\vspace{-0.25cm}
\textbf{Quantitative comparisons on the MPI LF dataset.}
Furthermore, we also evaluated the robustness of different methods under different disparity ranges on the MPI dataset \cite{Vamsi2017}. Each scene  contains a high-angular densely-sampled LF image composed of $101$ SAIs distributed on a scanline with spatial resolution $720\times 960$. The disparity between adjacent SAI is around $1$ pixel. We can sample SAIs with different intervals along the angular dimension to construct LFs with different disparity ranges. Specifically, we separately set $10$ disparity ranges from $8$ to $80$ pixels between two input SAIs. For each disparity range, we evenly sampled $3$ SAIs between two input SAIs as ground truth. The PSNR/SSIM shown in Fig.~\ref{fig:quanti_mpi} indicates that although the reconstruction quality of all methods decreases along with the increase of the disparity range, Ours consistently achieves better reconstruction quality than the other methods under all disparity ranges, demonstrating the robustness of our method towards different disparity ranges.

\begin{table}[!t]
\centering
\setlength{\tabcolsep}{0.9mm}
\caption{Comparisons of running time (in seconds per view) and model parameter size (M) of different methods on the Inria Sparse LF dataset \cite{shi2019framework}.}
\begin{tabular}{cccccc}
\toprule
                 & Baseline     & \makecell{Kalantari\\\textit{et al.} \cite{kalantari2016learning}}  & \makecell{Wu\\\textit{et al.} \cite{wu2019learning}} &\makecell{Jin\\\textit{et al.} \cite{jin2020deep}} & Ours   \\ 
\midrule
Time    &\textbf{0.65} &6.32                                                    &27.77                                    &0.69                                  &2.78\\
\# Params   &0.95          &2.55                                                    &\textbf{0.24}                            &2.22                                  &0.69\\
\bottomrule
\end{tabular}
\label{table:efficiency}
\vspace{-0.7cm}
\end{table}

\textbf{Qualitative comparisons}.
We visually compared reconstructed wide-baseline LFs by different methods. One of the challenges for the wide-baseline LF reconstruction is fairly reconstructing a large number of occlusion regions while are not so many in narrow-baseline ones. From reconstructed SAIs and zoomed-in regions shown in Fig.~\ref{fig:visual_inria} and Fig.~\ref{fig:visual_mpi}, it can be observed that our method can produce sharp edges at the occlusion boundaries, while other methods produce either severe distortions or heavy blurry effects at these regions. Moreover, our method can produce better high-frequency details at texture regions than other methods, which demonstrate that our method can still achieve high-quality LF reconstructions even under such large input disparity range. Please refer to the \textit{Supplementary Material} for more visual results.

\textbf{Comparisons of the LF parallax structure.}
Moreover, as the parallax structure is one of the most important values of LF data, 
we thus managed to compare the parallax structures of LFs reconstructed by different methods. On the one hand, comparisons of EPIs in Figs.~\ref{fig:visual_inria} and \ref{fig:visual_mpi} indicate that our method can preserve clearer linear structures than other methods, even for lines corresponding to large-disparity regions, validating the strong ability of our method in preserving parallax structure on extremely sparse LFs. On the other hand, it is expected that depth/disparity maps estimated from high-quality LFs shall be close to those are estimated from ground-truth ones. Thus, we compared the depth maps estimated from reconstructed LFs of different methods by a identical LF-based depth estimation method \cite{chen2018accurate}. As shown in Fig.~\ref{fig:visual_inria_depth}, our method can produce sharper edges at occlusion boundaries and preserve smoothness at regions with uniform depth, which are closest to the ground truth. Such observations also demonstrate the advantage of our method on preserving the LF parallax structure. 

\textbf{Efficiency comparisons}.
We compared the efficiency and model size of different methods.  All the methods were implemented on a Linux server with Intel CPU E5-2699 @ 2.20GHz, 128GB RAM and Tesla V100. As listed in Table~\ref{table:efficiency}, we can see that Ours is much faster than Kalantari \textit{et al.} \cite{kalantari2016learning}  and Wu \textit{et al.} \cite{wu2019learning} but slower than Baseline and Jin \textit{et al.} \cite{jin2020deep}. Besides, our model size is smaller than Baseline,  Kalantari \textit{et al.} \cite{kalantari2016learning}, and Jin \textit{et al.} \cite{jin2020deep}, but larger than Wu \textit{et al.} \cite{wu2019learning}. Taking the reconstruction accuracy, efficiency, and model size together, we believe our method is the best. 

\subsection{Ablation Study}
\begin{figure}[thp]
\begin{center}
  \includegraphics[width=\linewidth]{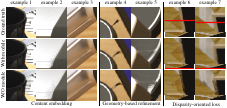}
\end{center}\vspace{-0.4cm}
  \caption{Effectiveness of the three key modules.}
\label{fig:visual_ablation}
\vspace{-0.4cm}
\end{figure}

\begin{figure}[thp]
\begin{center}
  \includegraphics[width=\linewidth]{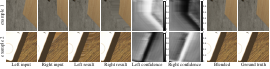}
\end{center}\vspace{-0.4cm}
  \caption{Effectiveness of confidence-based blending.}
\label{fig:visual_conf}
\vspace{-0.6cm}
\end{figure}

We carried out comprehensive ablation studies to validate the effectiveness of three key components involved in our framework, i.e., content embedding, the geometry-based refinement module, and the disparity-oriented loss term. Specifically, each component was sequentially added to the base model 
until all the three components were included to form the complete model. As shown in Table~\ref{table:ablation} and Fig.~\ref{fig:visual_ablation}, it can be seen that there is a significant increase of performance when adding the content embedding to the base model, which verifies the advantage brought by detecting the texture edges of input views, and understanding the occlusion and non-Lambertian relations between input views. The visual comparisons shown in Fig.~\ref{fig:visual_ablation} (examples 1-3) also verify the advantage.
Moreover, the geometry-based refinement module can also bring around $0.3$ dB increase of PSNR based on model with the content embedding. The examples 4-5 in Fig.~\ref{fig:visual_ablation} also shown that some fine structures such as delicate objects and textures are obviously broken without this module. It verifies the effectiveness of our refining pixel correlations in novel SAI through being guided by the correct ones from input SAIs.
By comparing results in the last two rows in Table~\ref{table:ablation} and the EPIs in examples 6-7 in Fig.~\ref{fig:visual_ablation}, we can see that supervising the parallax structure by the ground-truth disparity during  training is helpful to reconstruct high-quality LFs. We also provide the intermediate visual results before and after confidence-based blending in Fig.~\ref{fig:visual_conf}, where it can be observed that the confidence-based blending handles the occlusion regions by leveraging the advantages of the left and right results under the guidance of their confidence maps.

\begin{table}[!t]
\centering
\setlength{\tabcolsep}{0.8mm}
\caption{Ablation study. ``$\times$" denotes that the corresponding component is not included, while ``$\surd$"  denotes being included.}
\begin{tabular}{ccc|cc}
\toprule
\makecell{Content\\embedding} & \makecell{Geometry-\\based refinement} & \makecell{Disparity-\\oriented loss} &PSNR  &SSIM\\
\midrule
$\times$         &$\times$           &$\times$                   &29.05 &0.901\\
$\surd$      &$\times$          &$\times$                   &29.83 &0.915\\
$\surd$       &$\surd$       &$\times$                   &30.14 &\textbf{0.920}\\
$\surd$      &$\surd$        &$\surd$                &\textbf{30.33} &0.919\\
\bottomrule
\end{tabular}
\label{table:ablation}
\vspace{-0.25cm}
\end{table}

\section{Conclusion}
We have presented a learning-based method for densely-sampled LF reconstruction from extremely sparse ones.
More precisely, we focused on addressing the challenging problem of wide-baseline inputs 
and proposed a novel dynamic interpolation model.
By learning geometry-aware and content-adaptive interpolation weights via a lightweight neural network, 
our method overcomes the limitations of commonly-used warping operation, and 
efficiently reconstructs LFs with much higher quality, compared with state-of-the-art methods.
 
\clearpage
\balance
{\small
\bibliographystyle{ieee_fullname}
\bibliography{egbib}
}

\end{document}